\documentclass[letterpaper, 10 pt, conference]{ieeeconf}  

\IEEEoverridecommandlockouts                              

\usepackage{graphics} 
\usepackage{epsfig} 
\usepackage{amsmath} 
\usepackage{amssymb}  
\usepackage{subfigure}
\usepackage{comment}
\usepackage{color}

\title{\LARGE \bf
Drive Video Analysis for the Detection of Traffic Near-Miss Incidents
}

\author{Hirokatsu Kataoka$^{1}$, Teppei Suzuki$^{1,2}$, Shoko Oikawa$^{3}$, Yasuhiro Matsui$^{4}$ and Yutaka Satoh$^{1}$
\thanks{$^{1}$National Institute of Advanced Industrial Science and Technology (AIST)}
\thanks{$^{2}$Keio University}
\thanks{$^{3}$Tokyo Metropolitan University}
\thanks{$^{4}$National Traffic Science and Environment Laboratory (NTSEL)}
}

\begin{document}

\maketitle
\thispagestyle{empty}
\pagestyle{empty}

\begin{abstract}
Because of their recent introduction, self-driving cars and advanced driver assistance system (ADAS) equipped vehicles have had little opportunity to learn, the dangerous traffic (including near-miss incident) scenarios that provide normal drivers with strong motivation to drive safely. Accordingly, as a means of providing learning depth, this paper presents a novel traffic database that contains information on a large number of traffic near-miss incidents that were obtained by mounting driving recorders in more than 100 taxis over the course of a decade. The study makes the following two main contributions: (i) In order to assist automated systems in detecting near-miss incidents based on database instances, we created a large-scale traffic near-miss incident database (NIDB) that consists of video clip of dangerous events captured by monocular driving recorders. (ii) To illustrate the applicability of NIDB traffic near-miss incidents, we provide two primary database-related improvements: parameter fine-tuning using various near-miss scenes from NIDB, and foreground/background separation into motion representation. Then, using our new database in conjunction with a monocular driving recorder, we developed a near-miss recognition method that provides automated systems with a performance level that is comparable to a human-level understanding of near-miss incidents (64.5\% vs. 68.4\% at near-miss recognition, 61.3\% vs. 78.7\% at near-miss detection).
\end{abstract}

\section{Introduction}


With the emergence and innovative robotics technology, self-driving cars and advanced driver assistance system (ADAS) equipped vehicles, which can be seen as higher-level auto navigation robots, are rapidly being developed in the both academic and industrial fields. However, to date, such vehicle have had little opportunity to learn about, and thus recognize, the dangerous traffic (near-miss incident) scenarios that provide normal drivers with strong motivation to drive safely. 

\begin{figure}[t]
\centering
\subfigure[Close to a pedestrian]{\includegraphics[width=0.44\linewidth]{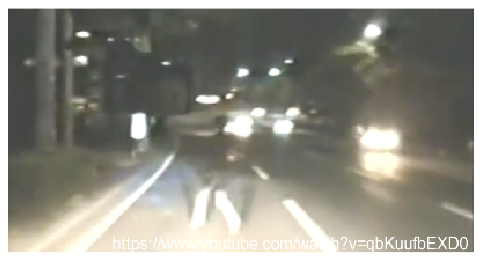}
\label{fig:topedestrian}}
\subfigure[Close to a vehicle]{\includegraphics[width=0.45\linewidth]{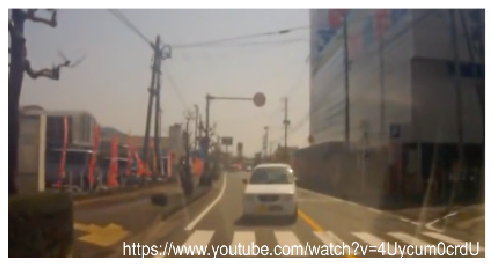}
\label{fig:tovehicle}}
\caption{Near-miss traffic incident scenes.}
\label{fig:nearmiss}
\end{figure}

Traffic near-miss incidents are dangerous situations in which collisions between vehicles and pedestrians or other vehicles had been narrowly avoided (see Figure~\ref{fig:nearmiss}), and the analysis of such incidents is an important step toward avoiding dangerous situations in self-driving vehicles. Existing traffic databases, such as the Caltech pedestrian dataset~\cite{DollarCVPR2009} and the Karlsruhe Institute of Technology and Toyota Technological Institute (KITTI) dataset~\cite{GeigerCVPR2012}, have increased in size over the past decade, but they do not include information on traffic near-miss incidents. This is important because the availability of a large corpus of near-miss traffic incident data, particularly if gathered from vehicle-mounted driving recorders, would provide researchers with greater opportunities to more deeply understand incident scenes. 

Although the collection of such data is very difficult due to the rarity of incidents in actual driving experiments, with the aim of improving the avoidance of such situations, we have collected a large-scale database containing videos of numerous traffic near-miss incidents. The analysis of such incident videos, however, is still challenging because most existing ways of representing motion are too ambiguous to capture the nuances contained in such scenes. For example, the left and right sides of Figure~\ref{fig:nearmiss} show urgent scenes, in which the vehicle-mounted drive recorders recorded near-miss proximity to a pedestrian and a vehicle, respectively. However, current motion representations may incorrectly label these incidents as \textit{crossing a street} (Figure~\ref{fig:topedestrian}) and \textit{driving straight} (Figure~\ref{fig:tovehicle}), even though they show potential impact dangers.

Here, there are two separate problems to be solved: (i) We must first collect and annotate a large-scale database of traffic near-miss incidents, and (ii) in order to perform a sophisticated analysis of near-miss incident scenes, we need to develop a way to accurately present such incident scenes in the database.


In this paper, we report on the development of a large-scale near-miss traffic incident database (NIDB)\footnote{The DB will be remained AIST private for a reason of copyright.} that contains a large number of annotated videos showing near-miss traffic incident scenes. The videos contained in this database are considered using two scenarios, the first evaluates near-miss traffic incident scenes from the perspective of driver feedback. The second considers temporal near-miss incident detection, including \textit{background} class, for self-driving and ADAS equipped vehicles. However, it is more difficult to detect traffic near-miss incidents when the setting contains a \textit{background} of ordinary traffic scenes. 

In summary, the contributions of our study are as follows:

\textbf{\underline{Conceptual contribution}:} Our philosophy is based on ``making sure the analysis of traffic near-miss incidents helps prevent collisions". To accomplish this, we have created a novel traffic database that contains videos of a large-number of traffic near-miss incidents in order to facilitate the analysis of such occurrences. When compared to existing databases such as the Caltech pedestrian~\cite{DollarCVPR2009} and KITTI~\cite{GeigerCVPR2012} datasets, our NIDB enables a more direct understanding of near-miss incidents.

\textbf{\underline{Technical contribution}:} To improve understanding of near-miss traffic incident scenes, we provide two primary improvements based on the NIDB: (i) training of traffic near-miss incident scenes, and (ii) extracting foregrounds from backgrounds using semantic flow. The results of our proposed NIDB-based data collection and video representation approach show that it produces a level of understanding regarding near-miss incidents that is close to human-level performance.


\section{Related work}

\subsection{Traffic data and approaches to its representation}

Several practical databases for pedestrian detection, such as the INRIA person dataset~\cite{DalalCVPR2005}, Caltech~\cite{DollarPAMI2012}, and the KITTI vision benchmark suite~\cite{GeigerCVPR2012}) have been proposed in the past decade.  The information contained in the KITTI database, which has been used to set meaningful vision problems for self-driving cars~\cite{GeigerCVPR2012} as well as problems related to stereo vision, optical flow, visual odometry, semantic segmentation, two- and three-dimensional (2D/3D) object detection, and 2D/3D tracking, has proven especially useful.

Thanks to the development of sophisticated approaches, such as fully convolutional networks (FCN)~\cite{LongCVPR2015} and region-based convolutional neural networks (R-CNN)~\cite{GirshickCVPR2014}, there has been improved performance of solving these problems using the KITTI. In addition, a manner of geometry  allows us to improve the rate of object detection~\cite{ChengCVPR2016} and optical flow~\cite{BaiECCV2016} not only in stereo~\cite{LuoCVPR2016}. As for semantic segmentation, we can now obtain knowledge about dense connections with graphical models and multi-scale CNN~\cite{KunduCVPR2016,ZhaoCVPR2017}. The usage of spatiotemporal analysis successfully predicts a future situation of pedestrians~\cite{KataokaBMVC2016,KataokaSensors2018}.

Unfortunately, none of these datasets contain scenes of near-miss incidents in which pedestrians, cyclists, or other vehicles must be avoided. Thus, there is an urgent need for a collection of incident scenes that can be used to train self-driving cars on how to safely navigate such dangerous situations.

\subsection{Video representations}

To date, space-time interest points (STIP) have been the primary focus for action recognition~\cite{LaptevIJCV2005}. In the STIP approach, the time $t$ space is added to the $x,y$ spatial domain. The most important aspect of this approach is that it uses dense trajectories (DT)~\cite{WangCVPR2011,WangIJCV2013} to track densely sampled feature points. In addition, Wang \textit{et al.} proposed improved dense trajectories (IDT)~\cite{WangICCV2013}, which estimates the camera motion in order to remove detection-based noise. This approach also incorporates a higher-order feature~\cite{KataokaACCV2014}.

Recently, temporal models with CNN have been proposed~\cite{TranICCV2015,SimonyanNIPS2014,WangCVPR2015}. 
For example, Tran~\cite{TranICCV2015} proposed a convolution model for xyt  maps that is based on the red-green-blue (RGB) sequence. The convolutional 3D (C3D) networks approach directly captures the temporal features contained in an image sequence. The recent investigations have revealed that the relationship between model depth and performance by using a 3D convolution~\cite{HaraCVPR2018,HaraICCVW2017}. Another approach, two-stream CNN, is a well-organized algorithm that captures the temporal feature of an image sequence~\cite{SimonyanNIPS2014}. 

The integration of the spatial and temporal streams allows us to effectively enhance the representation of motion, and thereby better understand how the spatial information relates to the temporal feature. Moreover, the strongest approach introduced thus far combines IDT and two-stream CNN. Trajectory-pooled Deep-convolutional Descriptors (TDD) have achieved a better level of performance for several benchmarks~\cite{WangCVPR2015}. The main idea behind this approach is to use improved trajectories to represent the convolutional maps extracted from the spatial and temporal streams.

While our approach primarily considered video representations, we believe that other approaches could be improved by additional NIDB training and semantic flow in relation to near-miss incident scenes.

 \begin{figure*}[t]
\centering

\subfigure[Number of videos per class: \textit{high-bicycle} (hi\_bic), \textit{high-pedestrian} (hi\_ped), \textit{high-vehicle} (hi\_veh), \textit{low-bicycle} (lw\_bic), \textit{low-pedestrian} (lw\_ped), \textit{low-vehicle} (lw\_veh), and \textit{background} (bg)]{\includegraphics[width=0.32\linewidth]{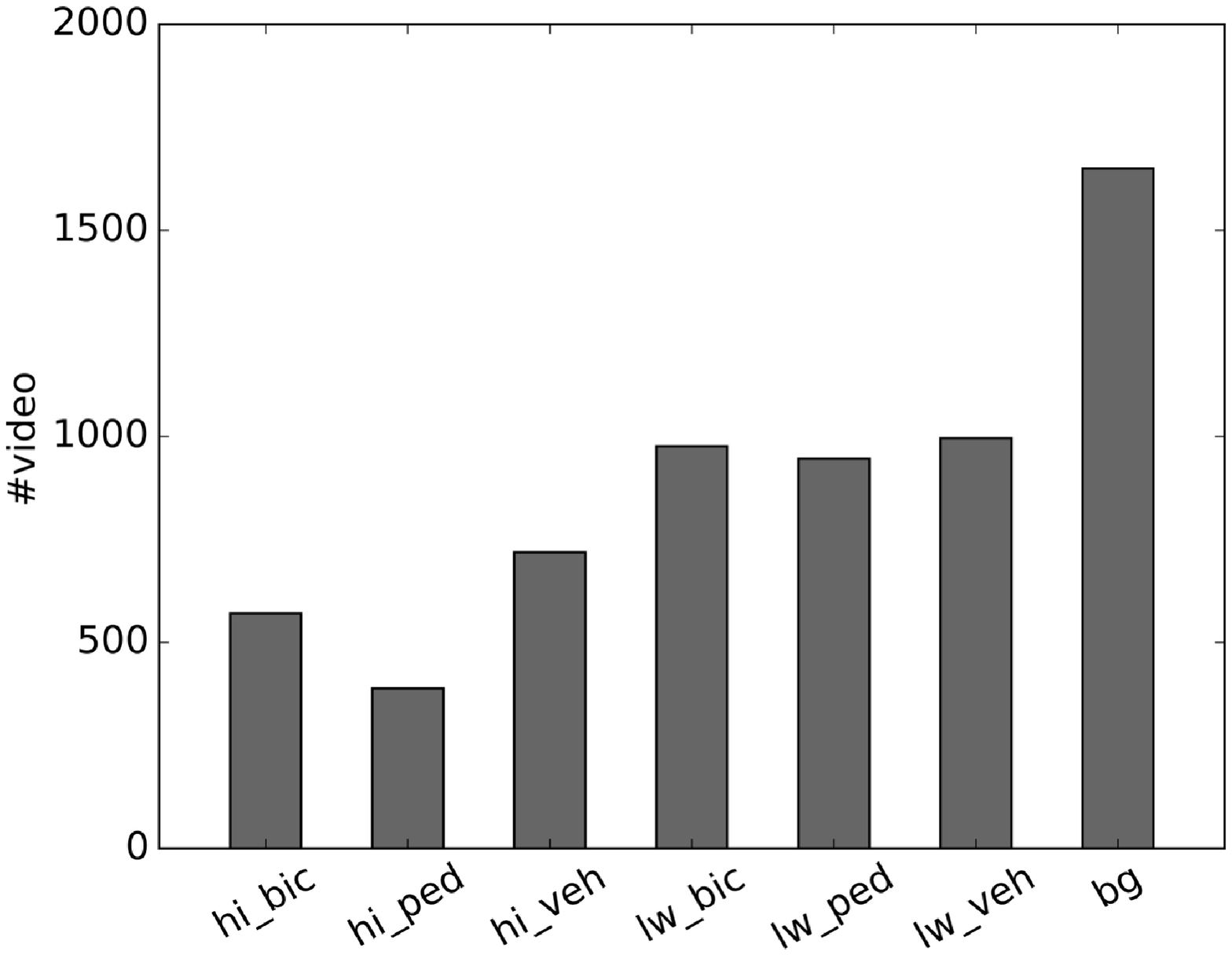}
\label{fig:sampleperclass}}
\subfigure[Traffic scene: parking (prk), cross road (cross), residential area (resid), main road (mainrd), and highway (high)]{\includegraphics[width=0.32\linewidth]{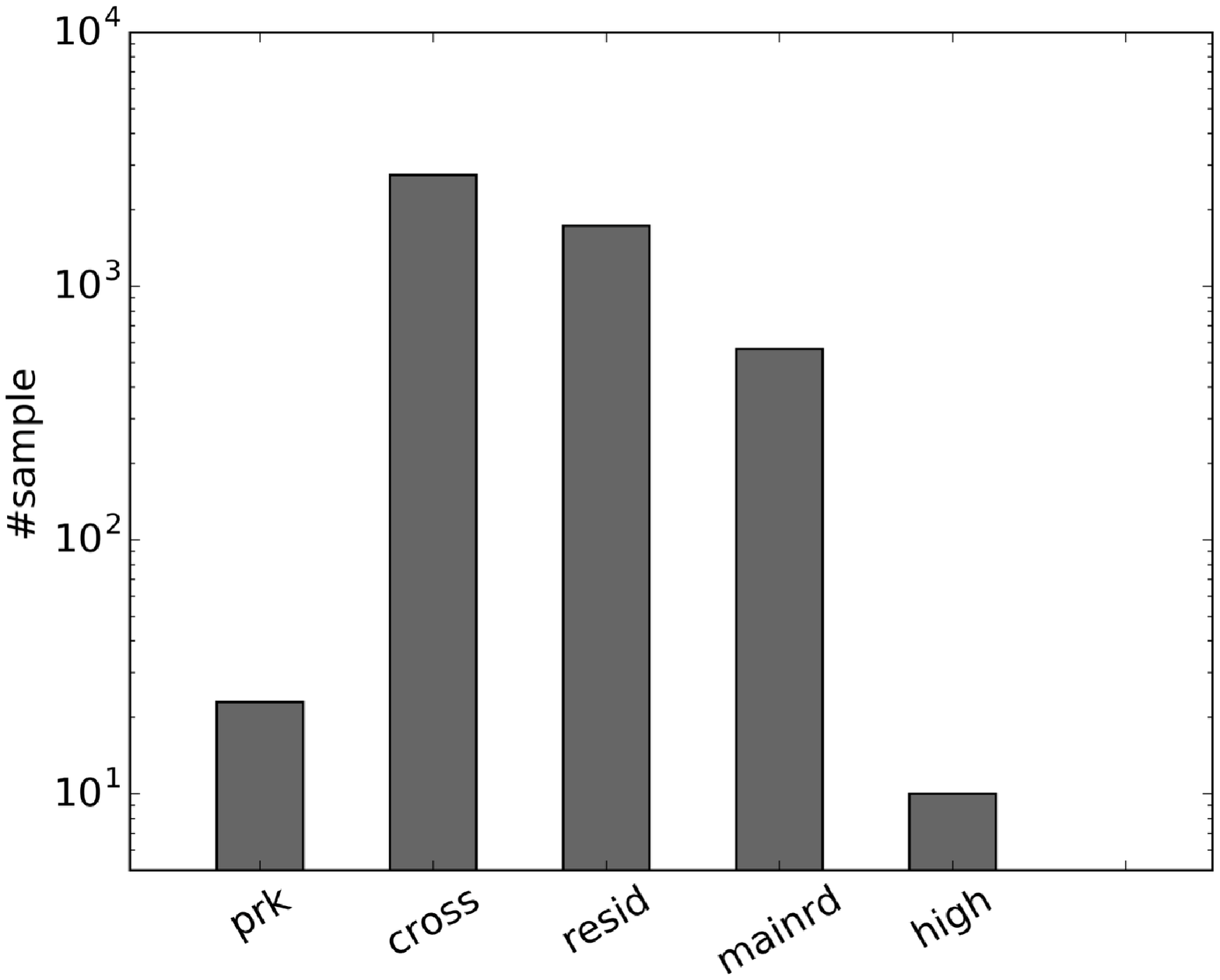}
\label{fig:trafficscene}}
\subfigure[Time period: day and night]{\includegraphics[width=0.32\linewidth]{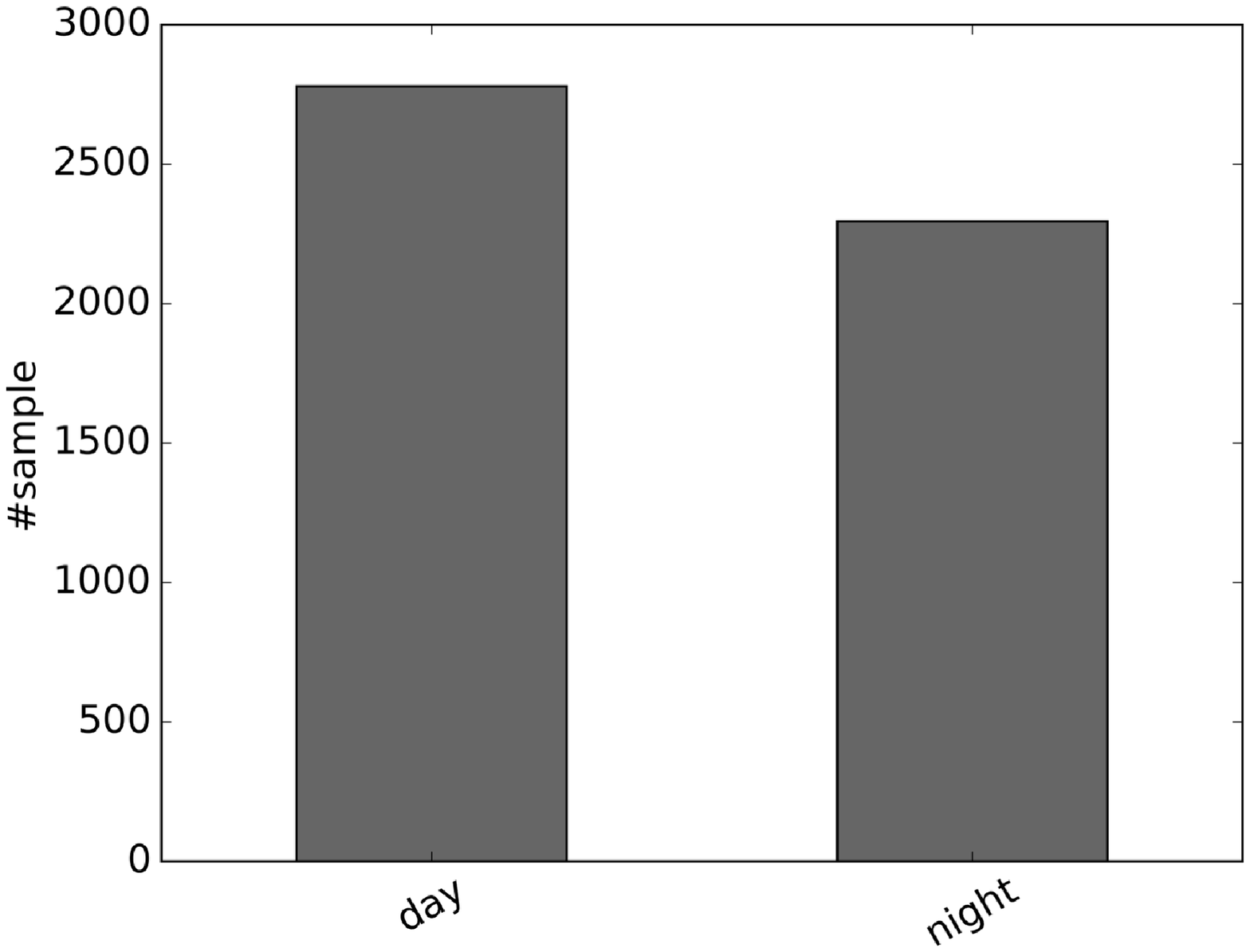}
\label{fig:timezone}}
\caption{Dataset statistics.}
\label{fig:nmdb_detail}
\end{figure*}

\section{Near-miss incident database (NIDB)}

In this section, we summarize the NIDB and discuss two scenarios for video classification tasks, database collection, annotation, and statistics. 

\subsection{NIDB Summary}
\label{sec:summary_nidb}
The NIDB provides video that can be used for better understanding the degree of danger and related elements. Overall, the database contains over 6.2 K videos and 1.3 M frames, many of which are incident scenes. The videos were captured using vehicle-mounted driving recorders. To the best of our knowledge, the NIDB is the first large-scale collection of videos depicting incident scenes.  The videos are divided into seven classes, including low/high risk for bicycles, pedestrians, and vehicles, as well as a background class. To these, we set the two following tasks:

\textbf{Near-miss incident recognition} consists of recognizing the following six classes of near-miss scenes in the observed videos: \{\textit{high-bicycle}, \textit{high-pedestrian}, \textit{high-vehicle}, \textit{low-bicycle}, \textit{low-pedestrian}, and \textit{low-vehicle}\}. The difference between high- and low-level danger is the proximity to collision and the driver's action. When the danger level is \textit{high}, the driver or safety system must react in such a way as to avoid an accident. However, when it is \textit{low}, the driver or safety system must simply be aware of the condition and be prepared for quick reaction. Therefore, it is necessary to clearly evaluate the hazard and danger level in the scene depicted in each image sequence. Note that the ability to recognize a near-miss situation can be used for insurance evaluations as well as for providing driver feedback.

\textbf{Temporal near-miss incident detection} consists of determining to which of the seven abovementioned classes (including \textit{background}) a scene belongs. The \textit{background} consists of scenes collected from driving records that do not include any hazards. The detection of near-miss scenes is difficult because, in addition to determining to which of the six near-miss categories the scene belongs, it is also necessary to recognize the difference between a dangerous scene and normal traffic. In a self-driving car, the primary focus of near-miss detection is aimed at avoiding such situations.

Since various traffic elements, such as bicycles, pedestrians, and other vehicles, appear in the background, it is necessary to have a vision system that can recognize a existence of dangerous conditions.

\subsection{Video collection, annotation and cross-validation}
\label{sec:annotation}

In this subsection, we describe the process of video collection (Step 1), annotation (Step 2), and cross-validation (Step 3). 

\subsubsection{Step 1: video collection}

Although it is difficult to collect near-miss videos, they are very beneficial for developing self-driving systems. In our database, videos were captured by mounting driving recorders in more than 100 taxis. These video recording systems were triggered to record for 15 seconds if there was sudden braking, resulting in deceleration of more than 0.5 G. Between 2006 and 2015, more than 60,000 videos were gathered. 

\subsubsection{Step 2: annotation}

We define traffic near-miss incidents and their low/high risks using the following annotation:

\begin{itemize}
\item Traffic near-miss incident definition: A traffic near-miss incident is an event in which an accident is avoided through driving operations such as braking and/or steering. Near-miss situations occur more frequently than collisions. In this paper, the proximity to collision of traffic near-miss incidents are extracted from the footage of video recorders mounted on taxis.
\item Low/high risk definition:
We evaluated (low/high) collision risk levels in situations where the driver did not take urgent actions such as emergency braking and/or steering operations. The high- and low-level danger categories correspond to the time-to-collision (TTC)~\cite{MatsuiTIP2013}. In case of a high-level risk, collision is imminent and the driver must react in less than 0.5 s (TTC $<$ 0.5s). For low-level risk, the TTC is more than 2.0 s (TTC $>$ 2.0s). Videos that show intermediate-level risk (0.5 s $\leq$ TTC $\leq$ 2.0 s), which is a mixture of high- and low-level risks, were not included in the NIDB because when training a convnet, it must be possible to make a clear visual distinction of risk. Accordingly, paper focuses solely low- and high-level risks in order to clearly divide the risk degree.
\end{itemize}

To avoid any ambiguity and strong bias in the data annotations, three expert annotators trimmed and categorized each of the videos based on the above definitions. Each video was assigned to a single category and was trimmed to a duration of 10--15 seconds. As the result of the annotation step, 60,000+ videos were selected, 5,000+ of which were near-miss incident videos.

\subsubsection{Step 3: Validation}

Validation was first conducted by the annotators, after which validators were tasked with improving the dataset annotation. In the validation step, it was necessary to process some operations such as annotation replacement (such as changing the risk-level from high to low) and video elimination (such as deleting unsuitable video). At the completion of this step, we had collected 4,594 near-miss incidents and an additional 1,650 background videos.

\begin{figure*}[t]
\begin{center}
   \includegraphics[width=0.90\linewidth]{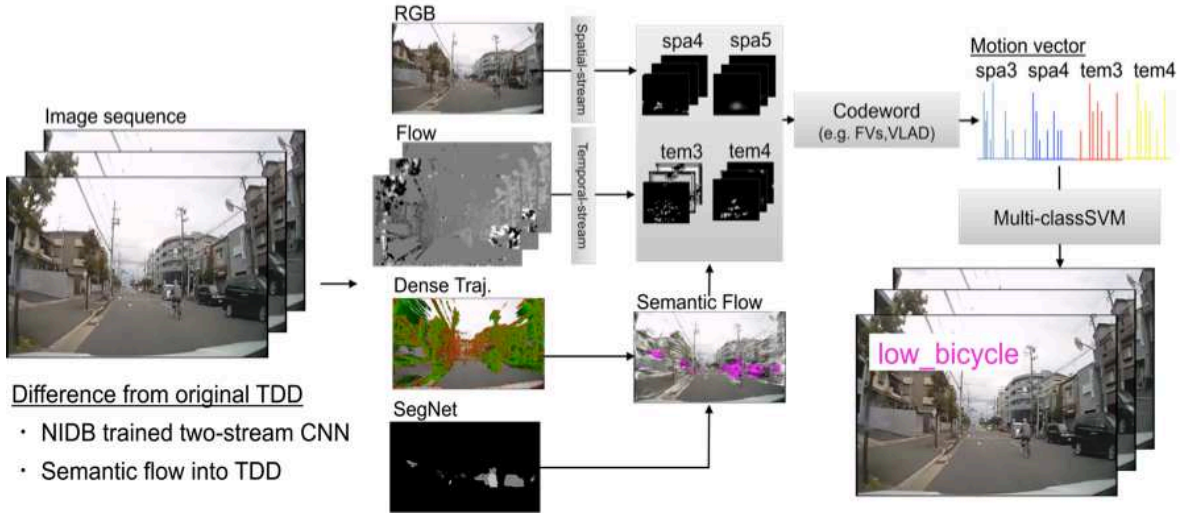}
\end{center}
   \caption{Flowchart of motion representation with NIDB.}
\label{fig:flowchart_stdd}
\end{figure*}

\subsection{Database analysis}

Figure~\ref{fig:nmdb_detail} shows the statistics for elements, danger level, traffic scene, time period, and weather. 

\textbf{Elements \& danger level (Figure~\ref{fig:sampleperclass}).} These categories were \{\textit{high-bicycle}, \textit{high-pedestrian}, \textit{high-vehicle}, \textit{low-bicycle}, \textit{low-pedestrian}, \textit{low-vehicle} and \textit{background}\}, and the \# of videos per class were \{570, 388, 718, 976, 946, 996, 1650\}.

\textbf{Traffic scene (Figure~\ref{fig:trafficscene}).} The traffic scenes were divided into five categories: parking, crossroad, residential area, main road, and highway.

\textbf{Time period (Figure~\ref{fig:timezone}).} The time periods were day and night (categorized based on sun brightness).


\section{Motion representation for understanding traffic near-miss incidents}

To detect traffic near-miss incidents, we applied TDD~\cite{WangCVPR2015} as a motion descriptor. Moreover, we provided two improvements: a pre-trained model with the NIDB (section \ref{sec:pretrain}), and semantic flow into the TDD (section \ref{sec:semanticflow}). In the near-miss incident scenes, we assumed that backgrounds would be noisy and that the near-miss objective (e.g. pedestrian) would be relatively small. Therefore, in order to efficiently describe a near-miss incident scene, it was necessary to create good features and separate the objective from the background. This sophisticated module is shown in Figure~\ref{fig:flowchart_stdd}, and a detailed description follows.

\subsection{Pre-trained model with the NIDB}
\label{sec:pretrain}

We started from a very deep two-stream CNN~\cite{WangarXiv2015} that was trained by the UCF-101 dataset. To fit the model into the NIDB, we executed two-step training. In the first step, we used only background videos without near-miss incidents in order to fit data from human actions to traffic-specified motions. In the second step, we trained near-miss incident videos in addition to the background videos.

After the two-step training, the NIDB pre-trained model was assigned in order to extract trajectory-based features from the convolutional maps in the TDD.

\subsection{Semantic flow into TDD}
\label{sec:semanticflow}

Next, to improve the TDD with semantic flow, we separated the foregrounds from backgrounds in each of the videos.

The TDD performs high-level video classification~\cite{WangCVPR2015} by combining hand-crafted IDT~\cite{WangICCV2013} with a deeply learned two-stream CNN~\cite{SimonyanNIPS2014}. The idea here is to access convolutional maps of spatial and temporal streams with a large number of improved trajectories. We confirmed that the combination of these methods resulted in enhanced performance. However, it remained difficult to extract useful features in near-miss scenes when the background was noisy or the motion was complicated. 

Recent studies verify the effectiveness of using semantic flow~\cite{LaraCVPR2016}, which is created by combining semantic segmentation and optical flow. Semantic flow is a concept that arises naturally in traffic safety and is used here when evaluating videos via the TDD. Next, semantic segmentation was implemented using SegNet~\cite{BadrinarayananarXiv2015}, which is a deep encoder-decoder architecture for multi-class pixelwise segmentation. In this step, the pretrained model of the  Cambridge-driving Labeled Video Database (CamVid) dataset was applied, but we used three objective categories (bicycle, pedestrian, and vehicle) when reconstructing the semantic flow. Following the improved trajectories~\cite{WangICCV2013}, we connected the Farnebäck optical flow. The semantic flow is calculated as follows:
\begin{eqnarray} 
   T^{'i}_{k} &=& T_{k}^{i} \ast S^{i}
\end{eqnarray}
where $T_{k}^{i}$ is the dense trajectory $k$ in the $i$th frame. $S$ is the result of semantic segmentation, where $S \in \{S_{bicycle}, S_{pedestrian}, S_{vehicle}\}$; and $T_{k}^{'i}$ is the semantic flow obtained by combining filtering $T_{k}^{i}$ and $S^{i}$, separated into foreground and background  ($T^{'} \in \{T_{fg}, T_{bg}\}$).

\begin{figure*}[t]
\centering
\subfigure[Original image.]{\includegraphics[width=0.32\linewidth]{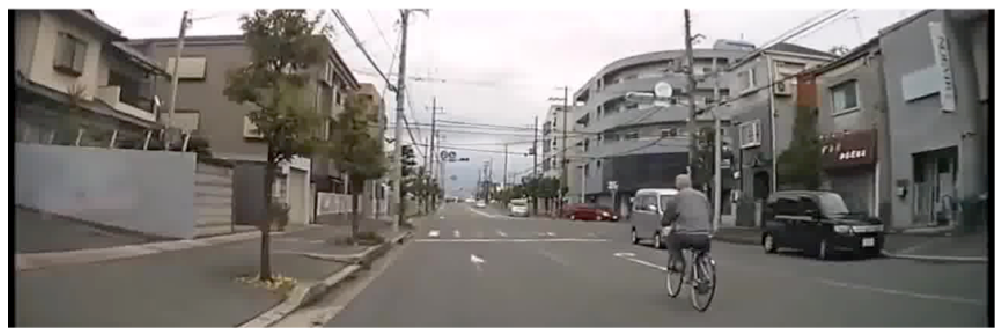}
\label{fig:sflow1}}
\subfigure[Semantic flow with fg (\textit{bicycle}, \textit{pedestrian}, and \textit{vehicle} are combined; magenta) and bg.]{\includegraphics[width=0.32\linewidth]{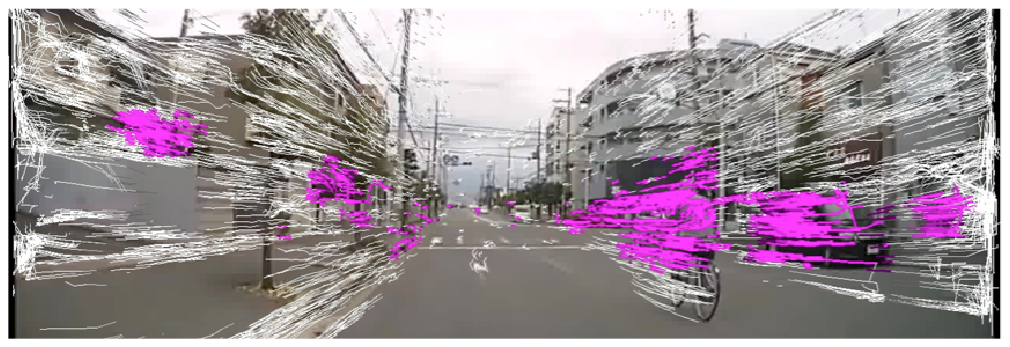}
\label{fig:sflow2}}
\subfigure[Semantic flow with separated semantics (\textit{bicycle}: yellow; \textit{pedestrian}: magenta; \textit{vehicle}: green) and bg.]{\includegraphics[width=0.32\linewidth]{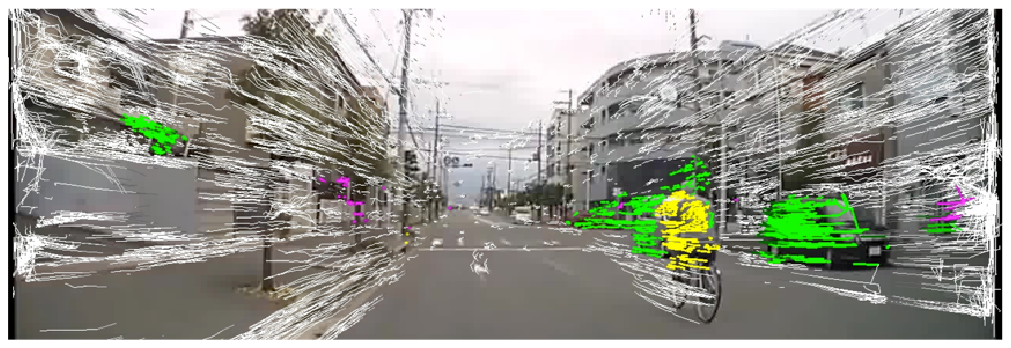}
\label{fig:sflow3}}
\caption{Semantic flow.}
\label{fig:sflow}
\end{figure*}

The semantic flow separates and combines the foreground/background effects. Figure~\ref{fig:sflow} shows an example of this process. When using filtered flows, it is only necessary to access the important elements in order to classify a scene. The length of each semantic flow is 15 frames, based on the IDT~\cite{WangICCV2013}.

To handle the neural network architecture, we used a base algorithm with the NIDB pre-trained very deep two-stream CNN~\cite{WangarXiv2015} in \ref{sec:pretrain}. To extract a feature, we assigned the fourth and fifth layers from the spatial stream, along with the third and fourth layers from the temporal stream, as in the original TDD~\cite{WangCVPR2015}. The TDD feature was thus captured as follows:
\begin{eqnarray}
 TDD(T_{k}^{'}, C_{m}^{a}) &=& \Sigma_{p=1}^{P} C_{m}^{a} ((r_{m} \times x_{p}^{k}), (r_{m} \times y_{p}^{k}), z_{p}^{k}) \nonumber \\
\end{eqnarray}
where $(x_{p}^{k}, y_{p}^{k}, z_{p}^{k})$ is the $p$th sampling point in the semantic flow $T_{k}^{'}$, and  $r_{m}$ is the $m$th scale ratio. Here, the dimensions of the TDD are 512-dim (spa4), 512-dim (spa5), 256-dim (tem3), and 512-dim (tem4). The TDD features are compressed as  64-dim to create a codeword for each convolutional map. Since a recent study~\cite{OhnishiACMMM2016} showed that use of Vectors of Locally Aggregated Descriptors (VLAD) is better than FVs for creating a TDD codeword vector (VLAD 92.0\% $>$ FV 91.3\%), a TDD-VLAD codeword vector format was adopted. We fixed the number of clusters to 256 and the dimension of the principal component analysis (PCA) to 64. The generated TDD-VLAD vector had a length of 16,384-dim.


\begin{table*}[t]
\begin{center}
\caption{Exploration study in our approaches: We showed (i) Our NIDB is effective to create a pretrained model (Ours0), (ii) the separation of fourground/background improves the accuracy (Ours1), (iii) IDT is beneficial for temporal detection (Ours2).}
\begin{tabular}{r|ccc|ccc}
 & & & & Ours0 & Ours1 & Ours2 \\
\hline
Spatial TDD & \checkmark & \checkmark & \checkmark & \checkmark & \checkmark & \checkmark \\
Temporal TDD & & \checkmark & \checkmark & \checkmark & \checkmark & \checkmark \\
Background fine-tuning & & & \checkmark & \checkmark & \checkmark & \checkmark \\
Near-miss fine-tuning & & & & \checkmark & \checkmark & \checkmark \\
Foreground \& background & & & & & \checkmark & \checkmark \\
Extra IDT features & & & & & & \checkmark \\
\hline
Recognition task & 56.3 & 58.6 & 60.0 & 63.2 & \textbf{64.5} & 62.1 \\
Temporal Detection task & 46.1 & 47.4 & 48.0 & 49.9 & 55.1 & \textbf{61.3} \\
\hline
\end{tabular}
\label{tab:improvements}
\end{center}
\end{table*}

\section{Experiments with NIDB}

Next, we considered two different tasks (near-miss recognition and temporal near-miss detection) on the NIDB. The near-miss recognition task contained six classes (\textit{high-bicycle}, \textit{high-pedestrian}, \textit{high-vehicle}, \textit{low-bicycle}, \textit{low-pedestrian}, and \textit{low-vehicle}). The temporal near-miss incident detection task included \textit{background} in addition to the six near-miss classes of the recognition task. The evaluation was based on one label per video, which is the same as in UCF-101~\cite{UCF101}. The train/test split of the NIDB is divided \{100, 50, 100, 100, 100, 100, 550\} (from \textit{high-bicycle} to \textit{background} in order) for the test, others for the train.

\subsection{Implementation details}

In the spatial stream, the input was 224 pixels $\times$ 224 pixels $\times$ three channels. The initial learning rate was set to 0.001, and updating was set to a factor of 0.1 per 10,000 iterations; thus, the learning was completed after 30,000 iterations. In the temporal stream, a basic stacked optical flow~\cite{SimonyanNIPS2014} was implemented in order to create an input of 224 pixels $\times$ 224 pixels $\times$ 20 channels. The initial learning rate was set to 0.001, and updating was set to a factor of 0.1 per 10,000 iterations; thus, the learning was completed after 50,000 iterations. We assigned a high dropout ratio in each of the fully connected (fc) layers, and set both the first and second fc layers to 0.9 for both streams.

\subsection{Parameter tuning}

We evaluated the following properties:

\textbf{Is the feature combination effective? (\textit{Spatial TDD} and \textit{Temporal TDD} in Table~\ref{tab:improvements}. The combined feature is better (+2.3\% recognition, +1.3\% temporal-detection), the temporal information is helpful for understanding a near-miss incident.)}: After confirming a certain percentage of the TDD on the both temporal-stream and spatial-stream modalities, we then captured a descriptor based on the original TDD that was obtained from the fourth and fifth layers in the spatial stream (spa4, spa5) and the third and fourth layers in the temporal stream (tem3, tem4). Table~\ref{tab:improvements} shows that when both features were combined, better results were obtained for both tasks. The effectiveness of adding the additional temporal-stream into the spatial-stream can be determined experimentally. Trajectories on the flow-based convolutional maps are helpful for interpreting near-miss situations. 


\textbf{Comparison of the fine-tuning with the NIDB only background, and fine-tuning with the NIDB including near-miss incidents (\textit{Background fine-tuning} and \textit{near-miss fine-tuning} (ours0) in Table~\ref{tab:improvements}. Background fine-tuning for +1.4\% recognition and +0.6\% temporal-detection, near-miss fine-tuning for +3.2\% recognition and +1.9\% temporal-detection)}: For the TDD, we employed very deep two-stream CNN~\cite{WangarXiv2015} in order to access the convolutional maps. Although the parameters in the baseline network were optimized with the UCF-101 dataset, we fine-tuned the NIDB, and then considered the NIDB both with just the background and after full fine-tuning. Table~\ref{tab:improvements} background and near-miss fine-tuning shows the results of these fine-tuning strategies. In this table, it can be seen that the NIDB with a fine-tuned background performed better (+1.4\% recognition and +0.6\% temporal-detection) when using the UCF-101 pretrained model. The convolutional maps were customized in order to focus on traffic scenes that contained images from the NIDB background. Moreover, when near-miss incidents were included, the recognition rates significantly increased (+3.2\% recognition and +1.9\% temporal-detection). These results show that it is important to fine-tune the near-miss videos in the CNN.

\textbf{Do we need to separate the vector with foreground/background? (\textit{foreground \& background} in Table~\ref{tab:improvements} (ours1); With foreground \& background, results are better-- +1.3\% recognition and +5.2\% temporal-detection, where the foreground includes three semantic meaning (bicycle, pedestrian, and vehicle)).}: When using the semantic meanings of the foreground, without a separated vector (the three categories were combined, see Figure~\ref{fig:sflow2}), the result is better than without foreground/background separation. We believe this is because the background includes various ego-motions that are relevant to the traffic scene. Moreover, we attempted to include recent semantic segmentation into the analysis (see Figure~\ref{fig:sflow3}), but the result was much worse than the results (ours1) shown in the table (51.1\% recognition and 37.0\% temporal-detection).

\begin{figure*}[t]
\centering
\subfigure[Spatio-temporal representations]{\includegraphics[width=0.42\linewidth]{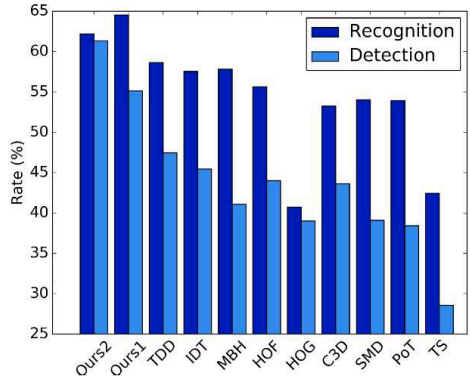}
\label{fig:sota_st}}
\subfigure[Spatial CNN models. A: AlexNet, V: VGGNet, fc: DeCAF (6 or 7 shows layer\#), I: ImageNet-(pre)train, N: NIDB-(pre)train]{\includegraphics[width=0.42\linewidth]{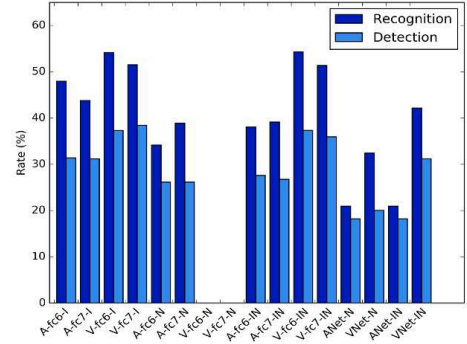}
\label{fig:sota_cnn}}
\caption{Comparison of state-of-the-art approaches: (a) spatio-temporal representations, (b) CNN models.}
\label{fig:param}
\end{figure*}

\textbf{\textit{Extra IDT features} in Table~\ref{tab:improvements}; -2.4\% recognition, +6.2\% temporal-detection.}: The combined approach (ours2) was significantly better at the temporal detection task (+6.2\%), but worse at the recognition task ($-$2.4\%) relative to ours1 without IDT. Use of the IDT results in a much better understanding of the background class, but the near-miss category is better understood with our TDD convolutional maps. The ours1 convolutional maps are  a concept-level descriptor used in NIDB learning.


\begin{figure*}[t]
\centering
\subfigure[Successful cases (from left to right): \textit{high\_vehicle}, \textit{low\_bicycle}, and \textit{high\_vehicle}]{\includegraphics[width=1.0\linewidth]{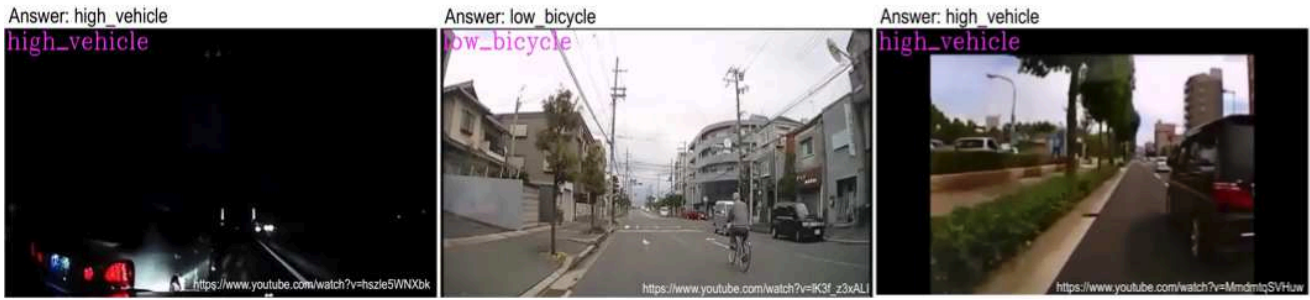}
\label{fig:success_nmdb}}
\\
\subfigure[Failed cases (from left to right): \textit{high\_bicycle}, \textit{high\_bicycle}, and \textit{background}]{\includegraphics[width=1.0\linewidth]{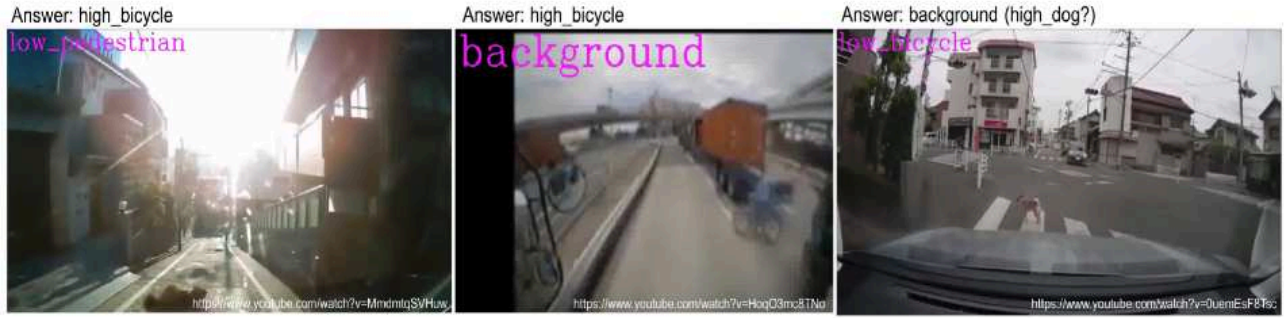}
\label{fig:failure_nmdb}}
\caption{Demonstrations.}
\label{fig:demo}
\end{figure*}

\begin{table}[t]
\begin{center}
\caption{Additional IDT features used with our motion representation and human-level recognition \& detection}
\begin{tabular}{ccc}
\hline
Representation & Recognition & Detection \\
\hline \hline
Ours1 & \textbf{64.5\%} & 55.1\% \\
Ours2 & 62.1\% & \textbf{61.3\%} \\
Human & \textbf{68.4\%} & \textbf{78.7\%} \\
\hline
\end{tabular}
\label{tab:extra}
\end{center}
\end{table}

Finally, we found that the performance rate of our model showed the most significant increase (+5.9\% recognition and +13.9\% temporal-detection) from the original TDD with the NIDB fine-tuning and semantic flow.

\subsection{Comparison}

Next, we compared our model with other state-of-the-art video representation models (see Figure~\ref{fig:sota_st}) and spatial features (see Figure~\ref{fig:sota_cnn}), as discussed below.

\textbf{DeCAF~\cite{DonahueICML2014} and end-to-end CNN~\cite{KrizhevskyNIPS2012,SimonyanICLR2015}.} 

Activation features and end-to-end models were extracted based on AlexNet~\cite{KrizhevskyNIPS2012} and VGGNet~\cite{SimonyanICLR2015}. In the Deep Convolution Activation Feature (DeCAF)~\cite{DonahueICML2014}, we set fc6 and fc7 for each CNN architecture. We used an ImageNet pretrained model (I), an NIDB pretrained model (N), and an ImageNet pretrained model fine-tuned with the NIDB (IN). The end-to-end models used the NIDB pretrained model (N) and the ImageNet used the pretrained model fine-tuned with NIDB (IN).

The video representations were as follows:

\textbf{IDT~\cite{WangICCV2013}.} The IDT is the de facto standard spatio-temporal model for hand-crafted video representation. The settings were based on the original implementation. To generate a codeword vector, motion boundary histograms (MBH) (192-d), histograms of optical flow (HoF) (108-d), and histogram of oriented gradients (HoG) (96-d) were captured at each trajectory sampling. The combined vector consisted of the MBH, HoF, and HoG features.

\textbf  {Pooled time series (PoT)~\cite{RyooCVPR2015} and subtle motion descriptor (SMD) ~\cite{KataokaBMVC2016}.} The settings for both were based on~\cite{KataokaBMVC2016}, which adjusted the parameters for short-term recognition from the understanding of a long-term event ~\cite{RyooCVPR2015}. We used a 10-frame accumulation to ensure high accuracy in near-miss recognition and detection. The multiclass classification was executed using a support vector machine (SVM). 

\textbf{C3D~\cite{TranICCV2015}.} The C3D networks employ a 3D convolutional filter on an xyt space that was obtained from the RGB sequence~\cite{TranICCV2015}. Fine-tuning is implemented by following the original C3D network with the NIDB. The number of iterations was set to 20,000.

Figures~\ref{fig:sota_st} and~\ref{fig:sota_cnn} compare the video representations and the CNN classifications, respectively.

Our model (64.5\% recognition, 61.3\% temporal-detection) significantly outperformed the other approaches for both NIDB tasks. For example, our results (+7.0\% recognition, +15.9\% temporal-detection) were better than those of the IDT (57.5\% recognition, 45.4\% temporal-detection). Here, both our model and the IDT relied on spatio-temporal features from improved trajectories. We can see the effectiveness of the deep-learned feature maps obtained by using the NIDB after fine-tuning and with various feature maps (spa4, spa5, tem3, tem4). 

The divided MBH, HOF, and HOG in the IDT also performed well when solving the two tasks. With IDT, the temporal MBH and HOF contributed to the discriminative descriptors. The SMD and PoT consisted of the differences between consecutive CNN activations and temporal pooling. The SMD was slightly better than the PoT for zero-around subtle motions, but the feature description from the entire image was redundant when attempting to understand a near-miss incident scene. Thus, it is important to carefully evaluate the dominant region around the relevant bicycle, pedestrian, or other vehicle in order to fully understand a near-miss scene. 

Although the trajectory-based approaches are discriminative, the two-stream CNN evaluates a feature using the entire image and has 42.4\% recognition and 28.3\% temporal-detection. In CNN-based classification, the fine-tuned VGGNet activation (fc6) gave the best performance on both tasks (54.3\% recognition, 37.3\% detection). When the ImageNet parameters were updated, they performed better than the NIDB pretrained model. Thus, it is important to use spatio-temporal models in order to obtain the best understanding of the NIDB scenes. 

Table~\ref{tab:extra} compares our approach and the combined our approach+IDT performance levels with that of humans. The human-level performance was 68.4\% recognition and 78.7\% detection. This shows the difficulty of analyzing a near-miss scene, even for humans. Our proposed method recognized near-miss scenes at a rate comparable to that of a human (human: 68.4\%, ours: 64.5\%). We note that this task is also difficult for humans. 

\subsection{Visual results}

We then evaluated our NIDB and pre-trained models using near-miss incident videos uploaded to video sharing service. Figure~\ref{fig:demo} shows successful examples of near-miss detection (Figure~\ref{fig:success_nmdb}) and failed detection (Figure~\ref{fig:failure_nmdb}) cases. The results show that our system outputs correct labels in most cases, but not always. Figure~\ref{fig:failure_nmdb} shows a missed object (bicycle), a very blurry image, and an unknown object (\textit{high\_dog}?).

\section{Conclusion}

In this paper, we presented a near-miss incident database (NIDB) that contains a large number of near-miss scenes obtained via vehicle-mounted driving recorders. The purpose of this database is to advance our understanding of near-miss scenes in order to improve safety systems for self-driving and ADAS-equipped vehicles. We also proposed TDD with semantic flow, which separates images into the foreground (near-miss objective) and background. The resulting successful near-miss incident data collection allows us to enhance the NIDB performance rate. 

For near-miss incident recognition, which involves categorizing near-miss incident scenes when there is no background (normal traffic scenes), the performance of our proposed method is close to that of humans (human: 68.4\%; our method: 64.5\%). For temporal near-miss incident detection, which is a joint classification problem for near-miss categories and backgrounds, our proposed method approaches that of humans (human: 78.7\%; our method: 61.3\%). Since it is easier for humans to classify background scenes, human temporal near-miss incident detection rates are higher than those for near-miss incident recognition. As an area of future work, we intend to improve the classification and traffic accident anticipation like~\cite{SuzukiCVPR2018}.

\bibliographystyle{IEEEtran}
\bibliography{nidb}

\end{document}